# Saliency Detection combining Multi-layer Integration algorithm with background prior and energy function

*Hanling Zhang;Chenxing Xia*

**Abstract**

In this paper, we propose an improved mechanism for saliency detection. Firstly,based on a neoteric background prior selecting four corners of an image as background,we use color and spatial contrast with each superpixel to obtain a salinecy map(CBP). Inspired by reverse-measurement methods to improve the accuracy of measurement in Engineering,we employ the Objectness labels as foreground prior based on part of information of CBP to construct a map(OFP).Further,an original energy function is applied to optimize both of them respectively and a single-layer saliency map(SLP)is formed by merging the above twos.Finally,to deal with the scale problem,we obtain our multi-layer map(MLP) by presenting an integration algorithm to take advantage of multiple saliency maps. Quantitative and qualitative experiments on three datasets demonstrate that our method performs favorably against the state-of-the-art algorithm.

# 1.Introduction

Salinecy detection aimed at identifing the most important and conspicuous object regions in an images has attracted much attention in recent years. There are various applications for salient object detection, including image segmentation[1],object recognition[2], image compression[3], image retrieval[4], dominant color detection[5] and so on.

Existing work of saliency detection can be roughly divided into two categories:top-down and bottom-up approaches.Top-down methods[6-8] are task-driven which generally require supervised learning with manually labeled ground truth.To better distinguish salient object from background,high-level information and supervised methods are incorporated to improve the accuracy of saliency map.The accurate of the method is high while the operator is complex and has a slow speed.On the other hand,bottom-up methods[9-11] usually exploit low-level cues such as features,color and spatial distances to construct saliency maps.Now more and more methods formulate their algorithms based on boundary prior,assuming that regions along the image boundary are more likely to be the background. However,it is not appropriate to put all nodes on the boundary into a class ,which will inevitably lead to noise.In addition,the accuracy of the saliency map is sensitive to the number of superpixels as salient objects are likely to appear at different scales.

However,most existing methods use single scale is not always the most optimal for different images which leads to miss many structural information.

The aforementioned problems are the breach of the paper.Firstly,we use color and spatial contrast with each superpixel to obtain a salinecy map(CBP) based on a neoteric prior selecting four corners of an image as background . Inspired by reverse-measurement methods to improve the accuracy of measurement in Engineering,hence,we put the investigative object to prospect which employs the Objectness labels as foreground prior to construct a map(OFP).Taking into account the respective insufficient, further,an original energy function is applied to optimize both of them respectively and a saliency map(SLP)is formed by combing the above twos .Finally,to deal with the scale problem,we generate multi-layer of superpixels with different granularities and present an integration algorithm to take advantage of multiple saliency maps(MLP).

In summary,the main contributions of our work include:

1)We proposed a novel background prior to construct a saliency map via an unique affinity which considers color and spatial contrast with each superpixel.

2)A novel energy function is put forward to optimize the background prior map and foreground prior map before incorporation.

3)Mutil-layer Integration algorithm is proposed to integrate multiple saliency maps into a more favorable result.

## 2.Related work

Recently, numerous bottom-up saliency detection methods have been proposed,which prefer to generate the saliency map by utilizing the boundary information.In[12],the contrast against image boundary is used as a new regional feature vector to characterize the background.In[14],a more robust boundary-based measure is proposed,which takes the spatial layout of image patches into consideration.[15] uses the four boundaries of an image as background cues to get foreground queries via manifold ranking(MR).[16] weights the initial prior map with boundary contrast to obtain the coarse saliency map.Inspired by previous academics,in this paper,we choose the features of four corners of an input image as background prior.

Generic object detection methods aim at generating the locations of all category independent objects in an image.We observe that object detection is closely related to saliency object segmentation.In[37],saliency is utilized as objectness measurement to generate object candidates.[38] uses a graphical model to exploit the relationship of objectness and saliency cues for saliency object detection. In[39], a random forest model is trained to predict the saliency score of an object candidate. In this work,we select the objectness labels as foreground prior .

Apart From detecting salient objects in a single layer,salient object detection also has been extended to identifying common salient objects shared in multiple layers.[22] concludes six principles for effective saliency computation and fuse them into a single framework via combining with Bayesian framework.[23] proposes multi-layer Cellular Automata to integrate multiple saliency maps into a better result under the Bayes framework.[24] obtains the final strong saliency map via the way of calculating the mean value.All of them achieve very good results demonstrating the effectivity of multi-layer in the accuracy of saliency detection.

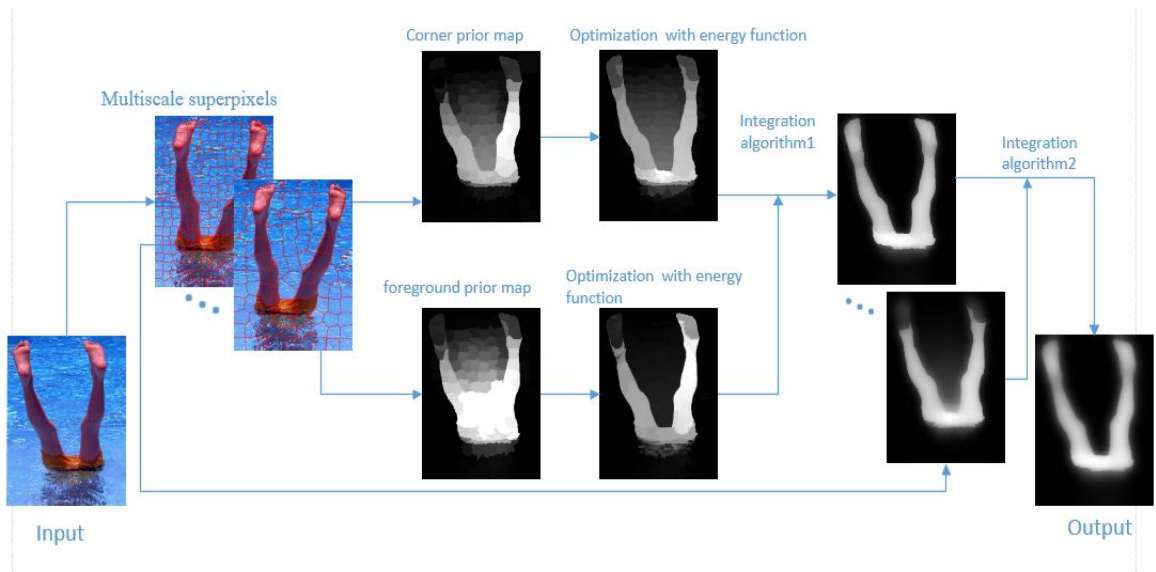

Fig.1.Pipeline of our algorithm.Firstly,two coarse saliency maps are constructed based on background prior with corner features and foreground prior with objectness label respectively. Then Combine the two saliency maps with intergration algorithm1 after optimized by a novel energy function separately.Finally,based on detection results at multiple scales,a strong saliency map is obtained by integration algorithm2.

### 3.Proposed Algorithm

Fig. 1 shows the main steps of the proposed salient object detection algorithm.In this section, we give the details about our model. To better capture intrinsic structure information and improve computational efficiency, an input image is over-segmented at M scales.At any scale m ,an image is segmented into N small superpixels by the simple linear iterative clustering(SLIC) algorithm[13].

**3.1** Select corner as Background prior(CBP)

In general, most people use four edges of the image as the background seeds and can also get good results.Howere,the method cannot deal well with the scene where some foreground noises may be in the border regions.This indicates that excessive use of the edge information as background prior will introduce noise.Based on the observation that the object occupies the four corners of the image seldomly, namely, the probability of the case that the object appears at the corners is less than borders while the four corners also includes background information. Hence,we extract the superpixels along the image the four corners of a image as prior background regions. Fig2 illustrates that this kind of background prior effect are also pretty good as the same with boundary prior in most cases and sometimes are better than the latter .

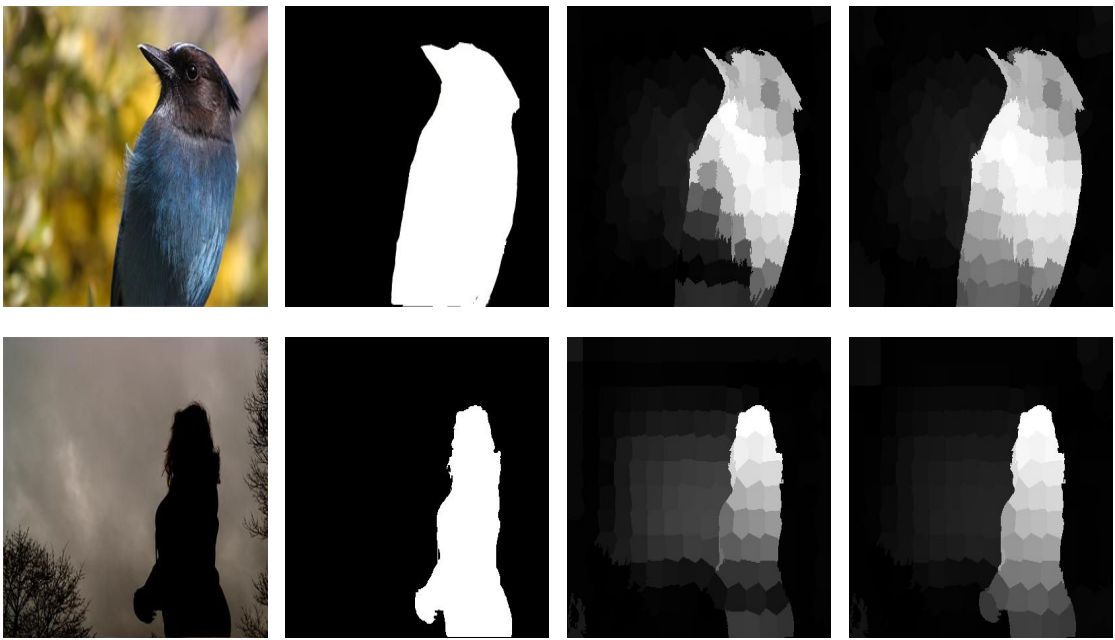

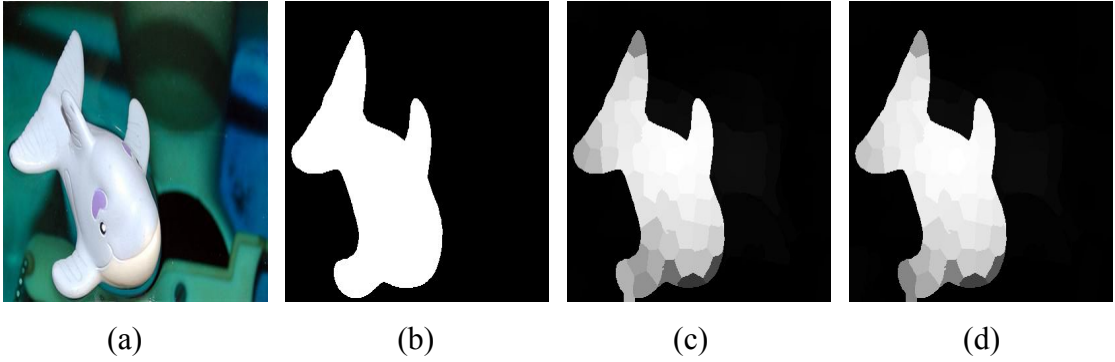

(a) (b) (c) (d)

Fig.2 Salient objects at different background prior.(a) Input image;(b) Ground truth;(c) Use boundary prior;(d) Use corner prior.

The similarity of two nodes is measured by considering their color and distance.Based on the intuition that neighboring regions are likely to share similar appearances and that the remote ones do not bother to have similar saliency values even if the appearance of them are highly identical.Our affinity between node i and node j is considered from the color and spatial characteristics.From the perspective of the color feature,we define the affinity entry $c_{ij}$ of node i to a certain node j as:

$$c_{ij} = \begin{cases} \exp(-\dfrac{\|c_i - c_j\|}{2\sigma_1^2}) & j \in N(i) \text{ or } i, j \in B \\ 0 & i = j \text{ or otherwise} \end{cases} \quad (1)$$

where $\|c_i - c_j\|$ is the Euclidean Distance between the node i and j in CIELAB color space. $\sigma_1$ (here is 0.1 )is a parameter controlling strength of the similarity.B denote the set of background, $N(i)$ indicates the set of the direct neighboring nodes of superpixel i,as well as the direct neighbors of those neighboring nodes.

Similarly,From the perspective of the spatial feature,we define the affinity entry $s_{ij}$ of node i to a certain node j as:

$$s_{ij} = \begin{cases} \exp(-\dfrac{\|\gamma_i - \gamma_j\|}{2\sigma_2^2}) & j \in N(i) \text{ or } i,j \in B \\ 0 & i = j \text{ or otherwise} \end{cases} \qquad (2)$$

where $\gamma_i, \gamma_j$ are the coordinates of the superpixel i and j.others abbreviation are same with Eq.(1).

Therefore,we define the affinity entry $w_{ij}$ of node i to a certain node j as:

$$w_{ij} = \begin{cases} c_{ij} \times s_{ij} & j \in N(i) \text{ or } i,j \in B \\ 0 & i = j \text{ or otherwise} \end{cases} \qquad (3)$$

where all abbreviation are same with Eq.(1) and Eq.(2).Therefore,we have an affinity matrix $W = [w_{ij}]$ is to indicate the similarity between any pair of nodes. The effects of affinity construction are illustrated in Fig.3.In order to normalize affinity matrix,a degree matrix $D = diag\{d_1, d_2, ..., d_N\}$ is generated, where $d_i = \sum_j w_{ij}$ .Finally,a row-normalized affinity matrix can be clearly calculated as follows:

$$G = D^{-1}.W \qquad (4)$$

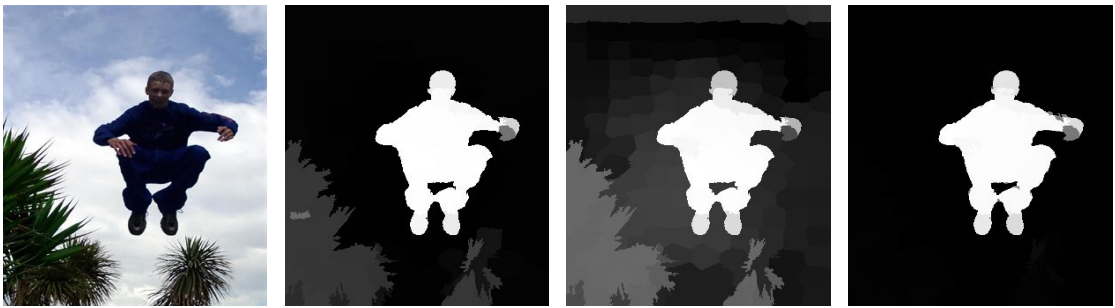

|  (a) | (b) | (c) | (d) |

Fig.3. Effects of affinity construction. (a) Input image;(b) Color contrast alone;(c) Spatial contrast alone;(d) Combine color contrast and spatial contrast.

We regrade the saliency of a superpixel as the contrasts to superpixels that on the four corners .In order to suppress the background effectively, we considers the four corners respectively. Let take the left-up corner of the image as an example, for a super-pixel i, based on the definition of affinity matrix, we define its saliency value $v_{lu}$ as:

$$v_{lu}(i) = (1 - \frac{1}{n}\sum_{j=1}^{n} g_{ij}) \times f(i) \qquad (5)$$

where $g_{ij}$ is the affinity entry defined in Eq.(4); n is the number of super-pixels on the left-up corner. In addition, $f(i)$ is the average value of the node i, computed by:

$$f(i) = \frac{1}{K_i} \sum_{q \in i} f(q) \qquad (6)$$

where $K_i$ indicates the number of pixels within region i.

Similarly, the saliency value of right-up corner, left-down corner and right-down corner can also be computed with the same method. Then we can obtain four maps: $v_{lu}, v_{ru}, v_{ld}$ and $v_{rd}$ respectively. Finally, we get a saliency map(CBP) by integrating the four maps according to the following equation:

$$v = v_{lu} \times v_{ru} \times v_{ld} \times v_{rd} \qquad (7)$$

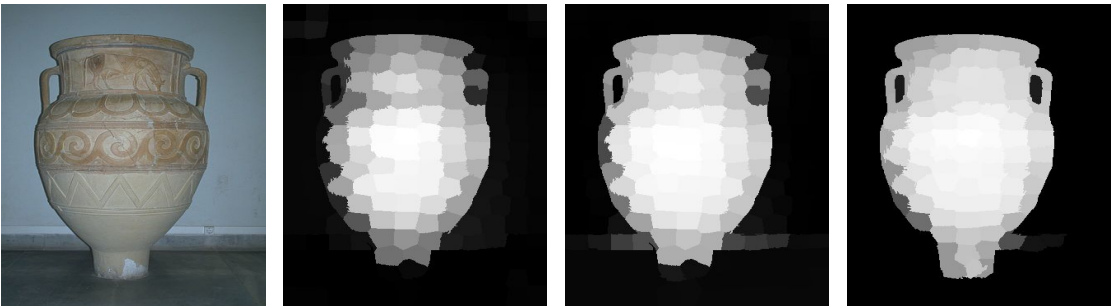

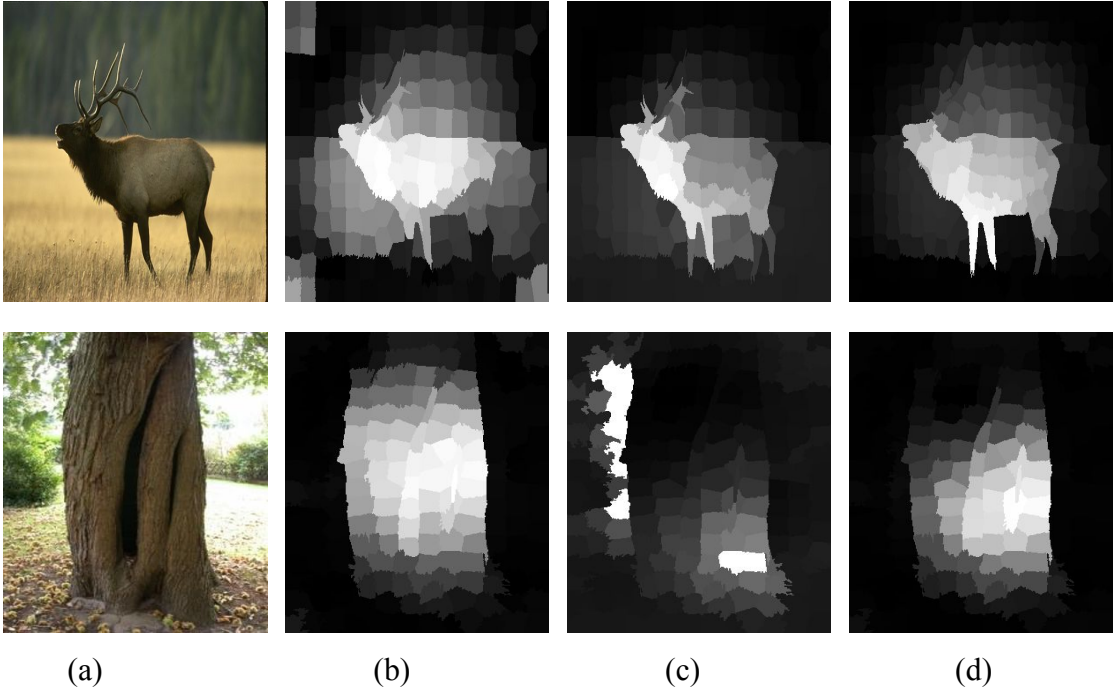

|  (a)  |  (b)  |  (c)  |  (d)  |

Fig.4. Objectness integration.(a) Input image;(b) Corner prior map;(c) Objectness prior map;(d)Combine corner with objectness prior without using energy function.

**3.2** Objectness labels as foreground prior(OFP)

As shown the middle row in Fig.4(b),however,depending on the corner prior alone might lead to high saliency assignment to the background regions.This promotes me to use some foreground prior to improve the result.In [25],an objectness score for any given image window is computed based on low-level cues.On the other hand on the other hand,Borji et al.[40] show that there is a center bias in some saliency detection datasets.On this basis,[26] proposes a Gaussian smoothing kernel of all sampling windowing to obtain the pixel-level objectness map $W(\tilde{p})$ via combining over-lapping scores :

$$W(\tilde{p}) = \sum_{h=1}^{H} P_h \cdot \exp[-(\frac{(x_p - \rho_x)^2}{2\sigma_x^2} + \frac{(y_p - \rho_y)^2}{2\sigma_y^2})] \qquad (8)$$

where $P_h$ is a probability score of the $h$-th sampling window. H is the number of sampling windows, $x_p, y_p, \rho_x, \rho_y$ denote the coordinates of pixel p and the center coordinates of window h respectively. Since saliency objects do not always appear at the image center as Fig5, the center-biased Gaussian model is not effectively and may include background pixels or miss the foreground regions. We use a model $W(p)$ with $\rho_x = x_o, \rho_y = y_o$, where $x_0, y_o$ denote the object region center of the CBP:

$$W(p) = \tilde{W}(p) * \psi_h \qquad (9)$$

where $\psi_h$ is the accuracy score, defined by:

$$\psi_h = \frac{\sum_{x_p, y_p} lab_h(x_p, y_p) \times v(x_p, y_p)}{\sum_{x_p, y_p} lab_h(x_p, y_p) + \beta} \qquad (10)$$

where $lab_h(x_p, y_p) = 1$ indicates that the pixel located at $(x_p, y_p)$ of the input image belongs to the h-th object candidate, and $lab_h(x_p, y_p) = 0$ otherwise; $v(x_p, y_p) \in [0,1]$ represents the CBP value of pixel $(x_p, y_p)$. Based on the pixel-level objecness map $W(p)$, we generate the region-level objectness map $W(i)$ which is the average of pixels' objectness values within a region:

$$W(i) = \frac{1}{n_i} \sum_{p \in i} W(p) \qquad (11)$$

where $n_i$ is the number of pixels in region i.

Based on the fact that high values of region-level objectness score calculated by Eq.11 can better indicate foreground areas. Fig.4(c) shows the saliency maps based on Objectness prior alone. The top and middle images effectively inhibit high values of the background saliency while the result of the bottom image is bad in some

scenarios.It indicates that it is a good choice to combine corner prior map with objectness prior map.

**3.3** Energy function and incorporation(SLP)

In this work, we propose a principled framework that intuitively integrates low level cues and directly aims for this goal. Since CBP and OFP are mingled with some noise,we optimize the two with energy function respectively before incorporation.The energy function is designed to assign the object region value 1 and the background region value 0,respectively.

At first,we binary the two maps with an adaptive threshold, $T_i$ denotes a certain node i :

$$T_i = \begin{cases} 1 & sal_i >= th \\ 0 & sal_i < th \end{cases} \quad (12)$$

where $sal_i$ is the value of node i. $th$ is an adaptive threshold,which we choose the mean of the map as the threshold[13],namely, $th = mean(sal)$.

Let $S_i$ is the saliency value of the superpixel i,the normalized $\xi(i)$ denotes the saliency value of each super-pixel in the above maps.Our energy function is thus defined as:

$$\arg\min \frac{1}{2}\left\{\sum_{i=1}^{N}[\log(1-\xi(i))]\cdot(s_i-1)^2 + \sum_{i=1}^{N}(1-T_i)\cdot s_i^2 + \sum_{i,j}w_{ij}\cdot(s_i-s_j)^2\right\} \quad (13)$$

The three terms define costs from different constraints.The first term encourages a superpixel i with large value $\xi(i)$ which is more likely to be foreground to take a large value $s_i$ (close to 1). Similarly, the second term encourages a node i with small value $T(i)$ which is more likely to be background to take a small value $s_i$ (close to 0). The last smoothness term encourages continuous saliency values. It indicates that a good saliency map should have similar saliency value between nearby super-pixels.

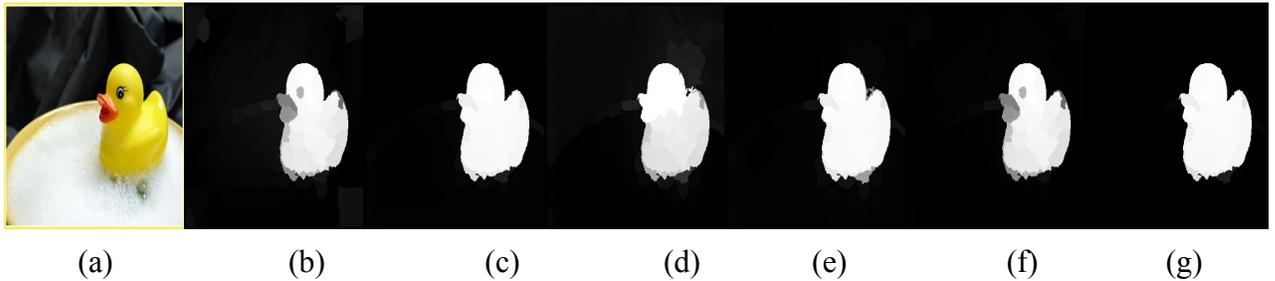

(a) (b) (c) (d) (e) (f) (g)

Fig.5. Optimization and Integration.(a) Input image;(b)Corner prior map without optimization; (c)Corner prior map after optimization;(d)Objectness prior map without optimization; (e)Objectness prior map after optimization;(f) Integrate map(b) and map(d);(g)Integrate map(c) and map(e).

The minimum solution is computed by setting the derivative of the above energy function to zero. The three items can achieve impressive results and the optimization can be done fast due to the small number of super-pixels.

Hence,we obtain two optimized saliency maps.Both of the maps are complementary to each other.The BCP can highlight the object more uniformly while the OFP can better suppress the background noise.Therefore,we incorporate them into a unified formula:

$$SLP = CBP \cdot ( 1 - \exp^{-\eta \cdot OFP} ) \quad (14)$$

where $\eta$ is a balance factor between them which is empirically set to 6 in our experiments. We observe that we can get a saliency map not only restraining the background but also highlighting the object.Fig.5(g) shows the effect of integration of two maps which are optimized by energy function.

### 3.4 Multi-layer Integration(MLP)

The accuracy of the saliency map is sensitive to the number of superpixels as salient objects are likely to appear at different scales. To deal with the scale problem, we generate mutil-layer of superpixels with different granularities,where N = 100, 150, 200, 250,300 respectively.Accordingly, we can get the same amount of saliency map.We represent the saliency map at each scale as $\{SLP^m\}$.In[4][5],their approaches is to strike the average as the final result, namely, $S = \frac{1}{M}\sum_{m=1}^{M} SLP^m$. Unlike what they do,we used a novel algorithm for integration.

At first,we calculate the similarity $SM_{ij}$ between two maps .The calculation of similarity is very simple, that is at the pixel-level, the similarity plus 1 when the corresponding coordinates has the same value of pixel, and then the cumulative result is normalized to [0,1],namely:

$$SM_{ij} = \frac{\sum_{x,y} SM_{ij}^{xy}}{m \times n} \quad (15)$$

where $x, y$ denote the coordinates of pixel p, $i, j$ indicate two saliency maps.

Because we choose five different scales, so we can get a similar matrix of 5x5 $[SM_{ij}]_{5\times 5}$.Then what we will do is to find out the map m which is the biggest difference with fours .So, an judgment vector $y = [y_1,...,y_M]^T$ comes into being,where

$y_m = \frac{1}{M} \sum_n SM_{mn}$. We find out the subscript $m = index[\min(y)]$. Hence, we define our finally saliency map as:

$$\text{MLP} = \sum_{m=1}^{M} \tilde{y}_m \cdot SLP^m \qquad (16)$$

where vector $\tilde{y}$ is same with vector $y$ except from the place $\tilde{y}_m = 1$. We take full advantage of the characteristics of the five maps by increasing the weight of the least similar map. Finally, we further refine the sliency map with the guided filter[27]. Fig.6(g-j) shows the saliency map with different algorithm via the way of intergation. Fig.6(g-i) shows that algorithm loses a lot of information in fusion while Fig.6(g) illustrates that our algorithm is of better robustness than other methods.

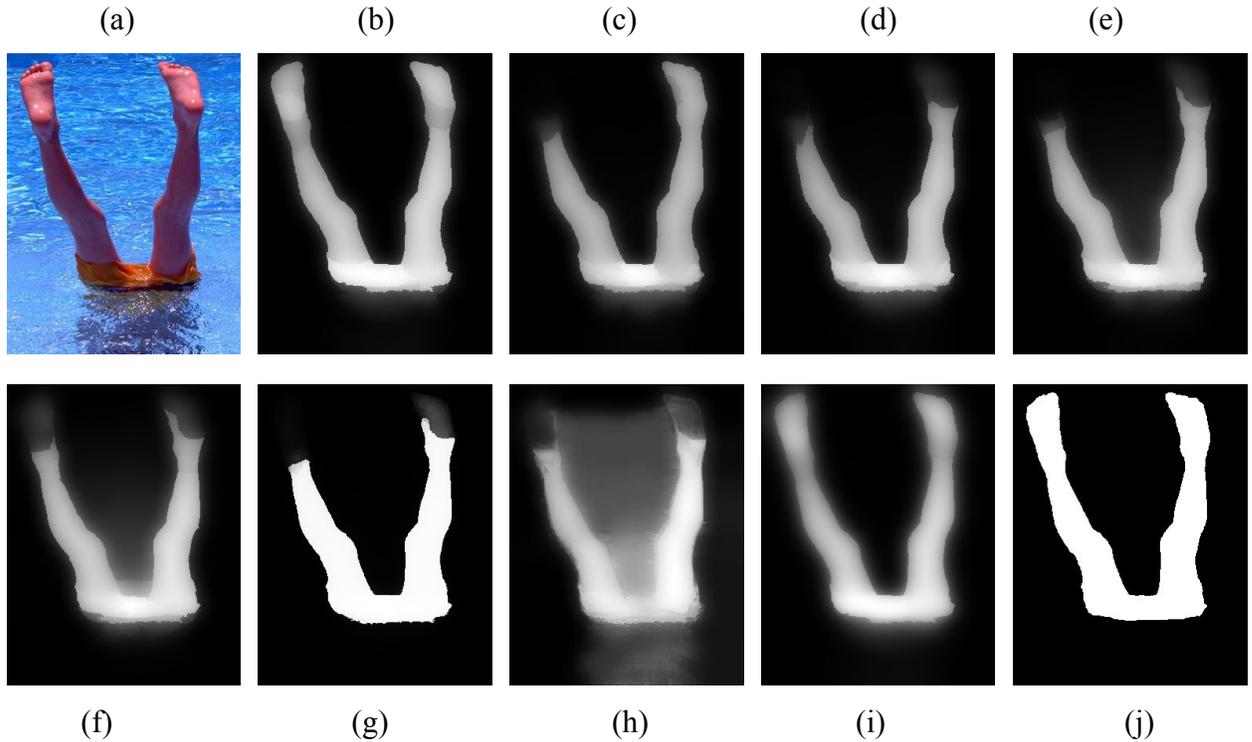

Fig.6. Effect of our algorithm via multi-layer integration. (a) Input image;(b-f)sliency maps with different superpixel N=100,150,200,250,300;(g)SCA algorithm[23];(h)MS algorithm[22];(i) Ours;(j)Ground truth.

## 4.Experiment Result

We evaluate the proposed method on the datasets: MSRA10K[35], SED1[34], and ASD[28].MSRA10K contains 10,000 randomly-chosen images from the MSRA dataset.SED1 contains 100 images of a single salient object annotated manually by three users.ASD consists of1000 images labeled with pixel-wise ground truth.

We compare our method with 6 state-of-the-art methods including the GS[29],SF[30],CP[31],PCA[32],CA[28],MB[33]on the MSRA10K,ASD and SED1 datasets.

### 3.1 Qualitative Results

We present some results of saliency maps generated by six methods for qualitative comparison in Fig.7.The result shows that the saliency maps generated by the proposed method highlight the saliency objects well with fewer noisy results. In the paper,while we choose the corner information as the background,the result is well.In addition,we utilize objectness labels as foreground prior to generate foreground map.The detected foreground and background in our maps are smooth due to the import of the energy function .At last,the novel multi-layer integration algorithm improves the effect further.

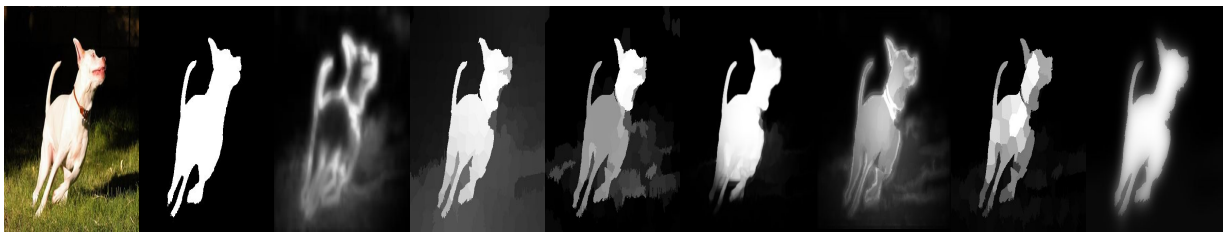

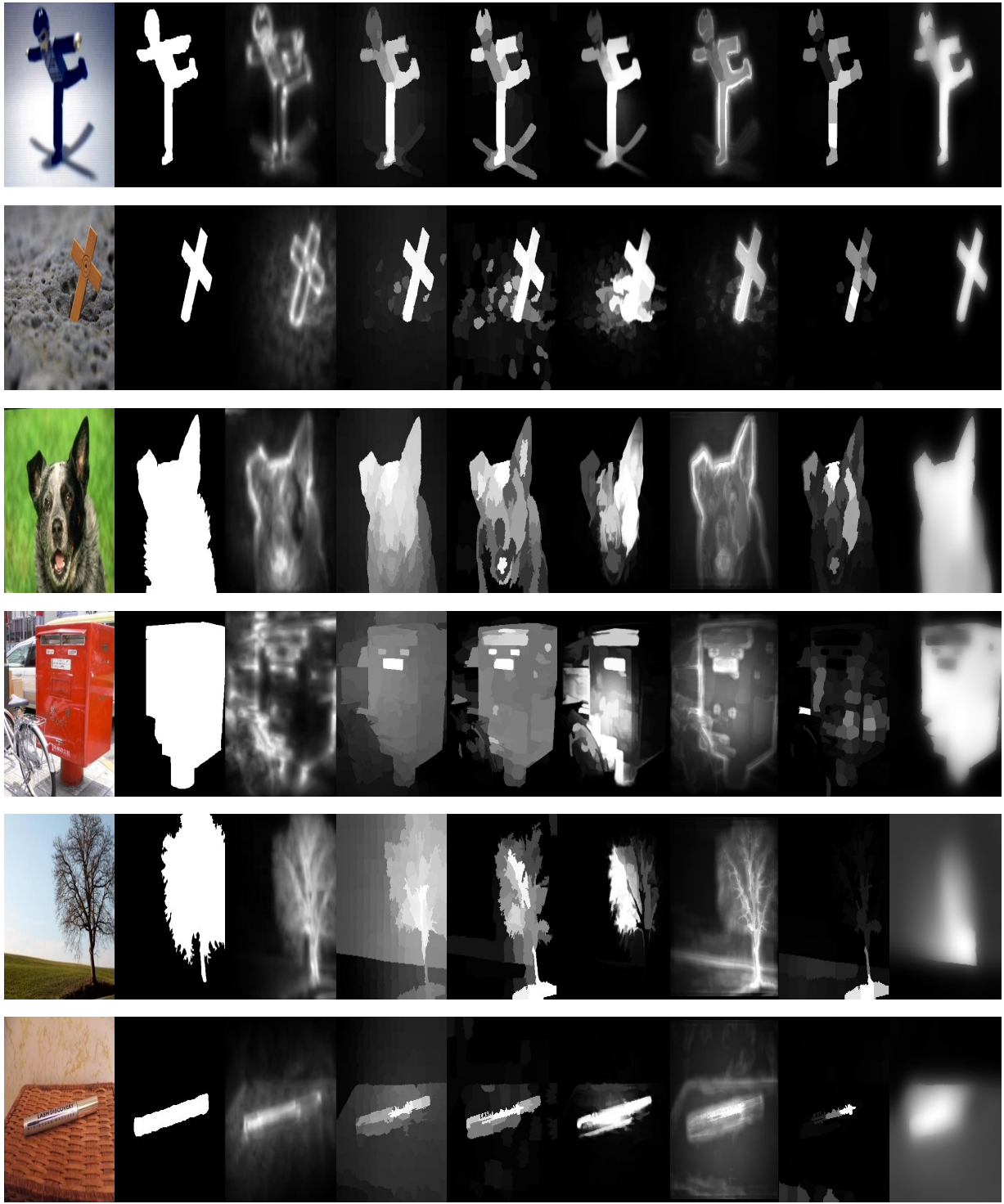

(a)　　　(b)　　　(c)　　　(d)　　　(e)　　　(f)　　　(g)　　　(h)　　　(i)

Fig.7. Qualitative comparisons of different approaches.(a)Input image;(b)Ground truth; (c)CA; (d)CP;(e)GS;(f)MB;(g)PCA;(h)SF;(i) Ours.The top two rows are examples in ASD,the middle row is example in MSRA and the bottom is in SED1.

## 3.2 Quantitative Results

We evaluate all methods by precision,recall and F-measure. The precious value represents the ratio of salient pixels correctly in all the identified pixels,while the recall value is indicated as the proportion of detected salient pixels corresponding to the ground-truth numbers. We obtain the precision-recall curves with binarizing the saliency map with a threshold sliding from 0 to 255. Fig.8(a1-a3) show the P-R curves where several state-of-the-art methods and the proposed algorithms perform well.

In addition, we measure the quality of the saliency maps using the F-Measure. The average precision and recall values are computed based on the generated binary masks and the ground truth while the F-Measure is computed by

$$F_{\beta} = \frac{(1+\beta^2)\text{Precision}\times\text{Recall}}{\beta^2 \times \text{Precision} + \text{Recall}} \qquad (16)$$

where we see $\beta^2$ to emphasize the precision[36].Fig.8 (b1-b3) show the F-Measure values of the evaluated methods on the three datasets. Overall,the proposed algorithms perform well.

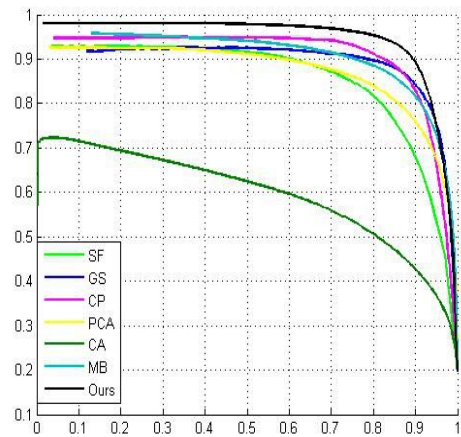 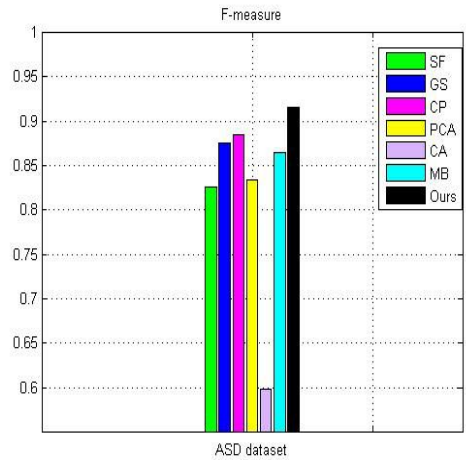

(a1) ASD P-R curve　　　　　　　　(b1) ASD F-measure

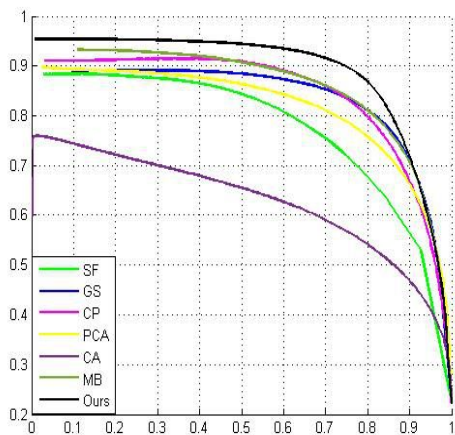 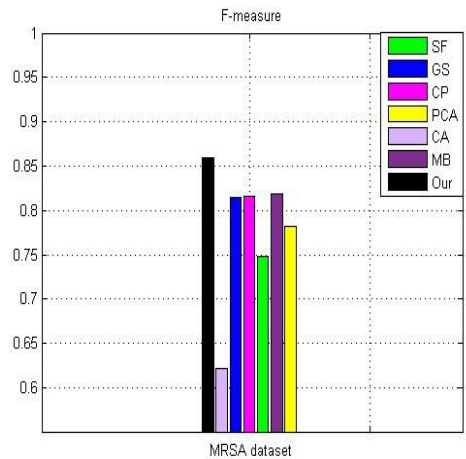

(a2) MSRA P-R curve　　　　　　　(b2) MSRA F-measure

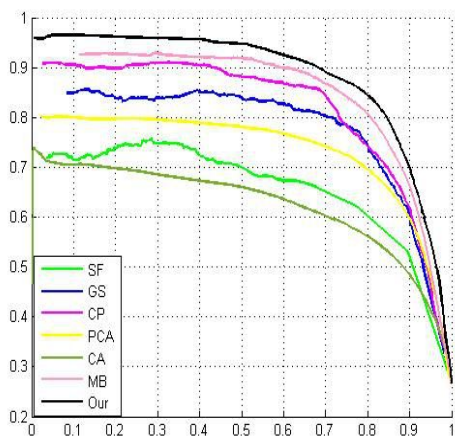 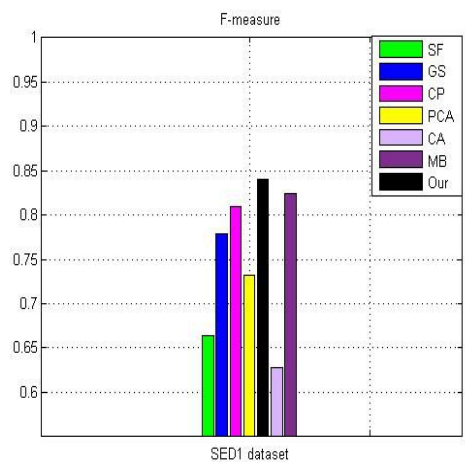

(a3) SED1 P-R curve　　　　　　　(b3) SED1 F-measure

Fig.8 Evaluation results on different datasets.From top to down:ASD ,MSRA,SED1.From left to right:the P-R curves and F-measure.

## 3.3 Limitation and Analysis

Our model performs favorably against existing algorithms with higher precision and recall.However,as the background prior map based on corners feathers which is insufficient in some scenarios and the foreground prior map based on the locations of object which may be unsafe if the positioning is not accurate,the proposed method does not work well if the first-stage is not well.The images in the last two rows of Fig.6 list failure cases of our saliency map models.However, we believe that investigating more sophisticated feature representations for our algorithm would be greatly beneficial. It would also be interesting to exploit top-down and category-independent semantic information to enhance the current results. We will leave these two directions as the starting point of our future research.

## 4 Conclusions

In this proposed a multi-layer integration for saliency detection based on selective background and objectness label.Firstly,based on a neoteric selective corner as background prior,we use color and spatial contrast with each superpixel to obtain a salinecy map(CBP).Then,we put the investigative object to prospect which employs the Objectness labels as foreground prior to construct a map(OFP).Taking into account the respective insufficient, an original energy function is applied to optimize both of them respectively,and a saliency map(SLP)is formed by merging the above twos .Finally,to deal with the scale problem,we generate multi-layer of superpixels with different granularities and present an integration algorithm to take advantage of multiple saliency maps(MLP). Experimental results demonstrate the effectiveness of our model.Our method achieves superior performance in terms of different evaluation metrics,compared with the state-of-arts on three benchmark image datasets.